\documentclass[sigconf]{acmart}

\usepackage{amsmath} %
\usepackage{bbold} %
\usepackage{euscript}
\usepackage{epsfig} %
\usepackage{graphicx}
\usepackage{listings} %
\usepackage{moreverb} %
\usepackage{color}
\usepackage{xspace}
\usepackage{natbib}
\usepackage{bbm}
\usepackage{multirow}
\usepackage{algorithm}
\usepackage{algorithmic}

\usepackage{subcaption}

\newcommand{\model}{\mathcal{M}}

\DeclareMathOperator*{\argmin}{argmin}
\DeclareMathOperator*{\topk}{TOP-K}

\usepackage{amsthm}
\theoremstyle{definition}

\theoremstyle{plain}

\title{Measuring Recommender System Effects with Simulated Users}

\author{Sirui Yao}
\authornote{Work completed while at Google.}
\email{ysirui@vt.edu}
\affiliation{%
\institution{Virginia Tech}
}

\author{Yoni Halpern}
\email{yhalpern@google.com}
\affiliation{%
\institution{Google}
}
\author{Nithum Thain}
\email{nthain@google.com}
\affiliation{%
\institution{Google}
}
\author{Xuezhi Wang}
\email{xuezhiw@google.com}
\affiliation{%
\institution{Google}
}
\author{Kang Lee}
\email{kanlig@google.com}
\affiliation{%
\institution{Google}
}
\author{Flavien Prost}
\email{fprost@google.com}
\affiliation{%
\institution{Google}
}
\author{Ed H. Chi}
\email{edchi@google.com}
\affiliation{%
\institution{Google}
}
\author{Jilin Chen}
\email{jilinc@google.com}
\affiliation{%
\institution{Google}
}
\author{Alex Beutel}
\email{alexbeutel@google.com}
\affiliation{%
\institution{Google}
}

\renewcommand{\shortauthors}{Yao, et al.}

\setcopyright{none}
\copyrightyear{2020}
\acmYear{2020}
\acmDOI{}

\acmConference[FATES on the Web 2020]{Second Workshop on Fairness, Accountability, Transparency, Ethics and Society on the Web}{20-24 April 2020}{Taipei, Taiwan}
\acmBooktitle{Second Workshop on Fairness, Accountability, Transparency, Ethics and Society on the Web}
\acmPrice{}
\acmISBN{}

\begin{document}
\begin{abstract}
Imagine a food recommender system---how would we check if it is \emph{causing} and fostering unhealthy eating habits or merely reflecting users' interests?
How much of a user's experience over time with a recommender  is caused by the recommender system's choices and biases, and how much is based on the user's preferences and biases? 
Popularity bias and filter bubbles are two of the most well-studied recommender system biases, but most of the prior research has focused on understanding the system behavior in a single recommendation step.  How do these biases interplay with user behavior, and what types of user experiences are created from repeated interactions?

In this work, we offer a simulation framework for measuring the impact of a recommender system under different types of user behavior.  Using this simulation framework, we can (a) isolate the effect of the recommender system from the user preferences, and (b) examine how the system performs not just on average for an ``average user'' but also the extreme experiences under atypical user behavior.
As part of the simulation framework, we propose a set of evaluation metrics over the simulations to understand the recommender system's behavior.  Finally, we present two empirical case studies---one on traditional collaborative filtering in MovieLens and one on a large-scale production recommender system---to understand how popularity bias manifests over time.
\end{abstract}

\maketitle
\footnotetext{N.B. This paper was originally presented with the title: ``Beyond Next Step Bias: Trajectory Simulation for Understanding Recommender System Behavior''.}

\begin{figure}[t]
\centering
\includegraphics[width=0.8\columnwidth]{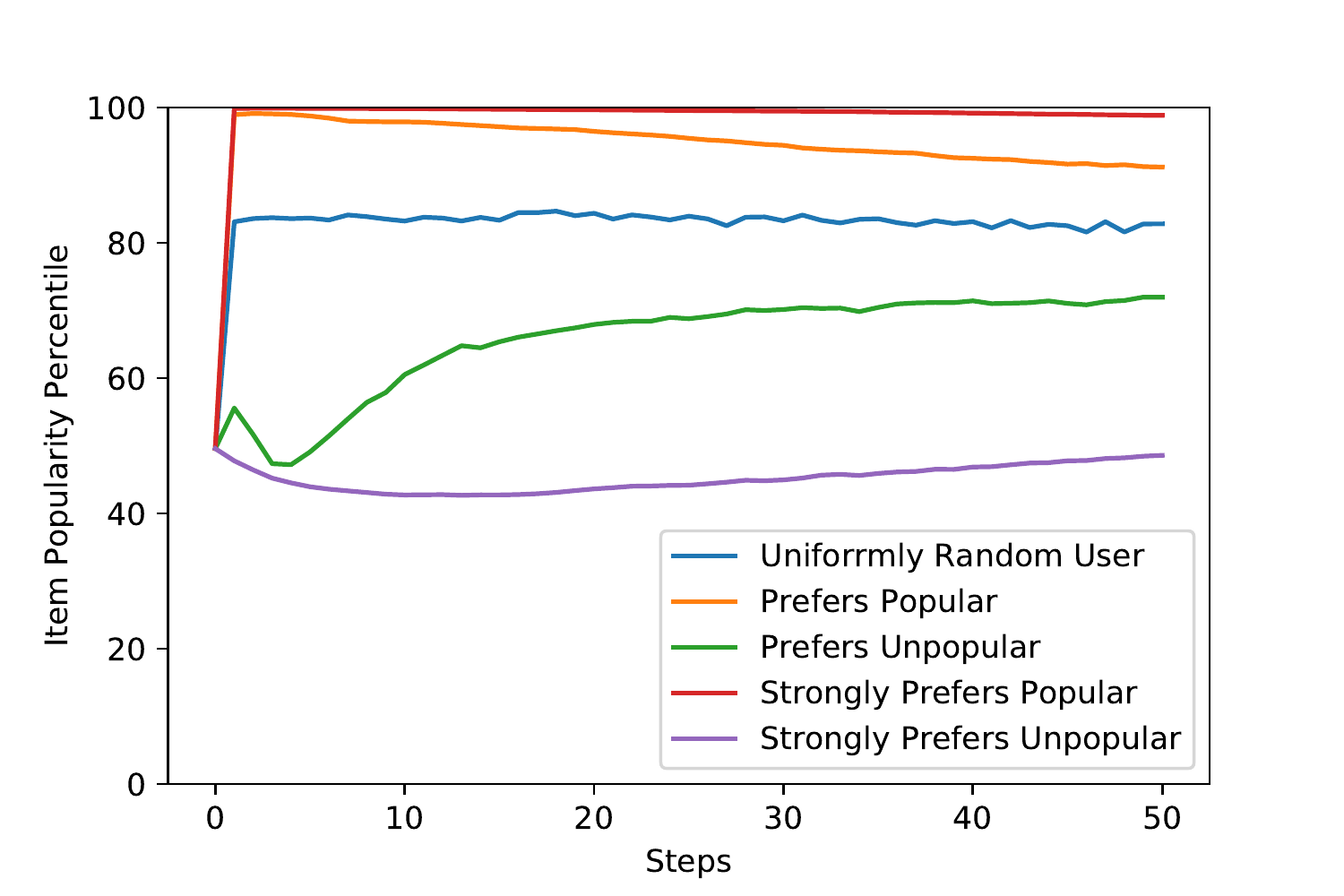}

\caption{Popularity bias is highly dynamic over time and depends significantly on user preferences as well. \label{fig:crownjewel}}
\end{figure}

\section{Introduction}
How can we measure if a food recommender system is \emph{causing} unhealthy eating habits by encouraging ``junk'' food and dessert or merely reflecting and reacting to users' interests?
More generally, what are the effects of a recommender system on a user's item consumption, and 
how does this evolve over repeated interactions between the user and the recommender?  How can we disentangle the effect of the recommender system's model, including its biases, and the effect of the user's preferences and biases?
Recommender systems are pervasive and have been studied for multiple decades now, yet there is still great debate about how to understand their behavior and measure their impact \cite{farzan2011encouraging, fleder2009blockbuster, beam2014automating, hasan2018excessive}.

Over the past decade, recommender systems have significantly increased in complexity, growing from collaborative filtering \cite{ekstrand2011collaborative} such as matrix factorization \cite{koren2009matrix} and item-to-item \cite{sarwar2001item} recommendation, to deep neural networks \cite{covington}.  Further, there has been a growing trend to model the temporal patterns of user behavior, using recurrent neural networks (RNNs) and attention models to capture past behavior \cite{rrn,hidasi2015session,xiao2017attentional} and reinforcement learning (RL) to optimize for future sequential actions \cite{Chen:2019:TOC:3289600.3290999,zheng2018drn,bouneffouf2012contextual}.  

While these developments have had a large impact in practical applications, the research community's understanding of them has comparatively lagged behind. Traditional recommendation evaluation focused on randomly holding out a fraction of user interactions with the system \cite{harper2016movielens,bennett2007netflix}.  This has more recently shifted toward using temporal splits to measure extrapolation to the future rather than interpolating in the past \cite{rrn}, and significant further effort has gone into accounting for sampling bias in the evaluation data \cite{schnabel2015unbiased,strehl2010learning}.
There has been significant controversy over whether neural networks are improving the state-of-the-art, in part because of the limitations of evaluation datasets \cite{dacrema2019we}.  However, all of these approaches are still primarily understanding the overall average accuracy of predicting the next step of user behavior.

Even as we have improved our understanding of accuracy, our understanding of the system's biases and the effect of those biases has been even less well-studied.  Likely the most well-studied recommender system bias is popularity bias, e.g., do recommenders over-recommend the most popular, i.e., head, items \cite{beutel2017beyond,Bellogin2017, DBLP:conf/flairs/AbdollahpouriBM19, channamsetty2017recommender}?  More recently, a broader set of concerns on fairness and bias in recommendation has been explored \cite{channamsetty2017recommender,DBLP:conf/kdd/SinghJ18,DBLP:conf/kdd/BeutelCDQWWHZHC19}. Likewise, researchers have studied diversity and ``filter bubbles'' to understand if users receive topically or politically narrow recommendations~\cite{bakshy2015exposure,dean2020recommendations} or are pushed towards ``radicalizing'' content~\citep{ledwich2019algorithmic,ribeiro2020auditing,munger2019supply}. These are multi-faceted, complex questions, where consensus on how to measure or address these challenges have hardly been reached.
 
In all of these investigations, even when considering the most well-studied systemic biases like popularity bias, existing evaluation frameworks fall short in answering important questions such as: Does the recommender system actually encourage users to consume popular content? How strong is this encouragement? If users are being encouraged to consume popular content, does it happen quickly or slowly? How varied is this experience across users? How sensitive are the resulting recommendations to a user's own preferences? 
Do new users with relatively short histories have a very different experience from users that have been interacting for longer?

These questions are difficult to address from observational data since the user's preferences (which are unobserved) interplay with the recommender system, and isolating the effects of the recommender itself is difficult.  Secondly, systemic biases can play out over multiple interactions, and a system's behavior may evolve over time. For example, Figure~\ref{fig:crownjewel} demonstrates how simulated users receive different amounts of popular content over the course of a simulation. In this paper, we focus on developing new tools and techniques to understand these issues and measure the behavior of recommender systems. 
We present a simulation framework for sequential interactions between users and a recommender system, and use this simulator as a controlled environment to collect synthetic trajectories of test users.

\vspace{2mm}
\noindent \textbf{Contributions.} 
We will describe the following contributions:
\begin{itemize}
    \item \textbf{Simulation Framework:} We offer our simulation framework and the relevant design decisions for effective measurement in \S \ref{sec:simulation_framework}.
    \item \textbf{Design of test users:} In \S \ref{sec:user_models}, we characterize multiple simulated test user behaviors and how they isolate different aspects of the recommender's impact.
    \item \textbf{Empirical Case Studies:} The simulation framework is general purpose. In \S \ref{sec:experiments} we use it to study popularity bias, a well-known property of many recommender systems. We analyze both a matrix-factorization based recommender trained on the public MovieLens dataset as well as a large-scale, production recommender system. This analysis yields a more detailed view of how popularity bias is reinforced by the long-term behaviors of recommender systems.
\end{itemize}

\emph{Relationship to observational metrics:} We believe this simulation framework provides complementary value to traditional observational metrics, such as accuracy \cite{harper2016movielens,bennett2007netflix} or even the large body of offline RL evaluation literature \cite{schnabel2015unbiased,strehl2010learning}.  In contrast, we explore how simulations can (a) isolate the recommender system's impact and (b) audit the recommender for unlikely, but still possible, negative user experiences.  For example, building on the food recommender example from above, even if only a few users are susceptible to skipping meals for dessert, it is valuable to know if the recommender system would encourage or reinforce that behavior, even rarely. Probing the recommender to discover potential consequences of recommendations on user experiences allows us to move beyond reasoning about the average user experience to understanding how user and system biases interact.

\section{Related Work}

\paragraph{Responsible Recommendation}
There has been an increasing interest in building more responsible recommendation systems.
\citet{bountouridis2019siren} proposes a simulation framework to visualize and analyze the effect of recommender systems in online news environments. \citet{nguyen2014exploring} studies the filter bubble effect in terms of content diversity and finds that recommender systems could lead users to a narrowing set of items over time. \citet{dean2020recommendations} audits recommender systems through the lens of user recourse. \citet{DBLP:conf/flairs/AbdollahpouriBM19} identifies and analyzes specific statistical biases (sparsity and popularity biases) and proposes methods that effectively neutralize them to a large extent. More recently, there has been an increased focus on fairness in recommendation as well \cite{singh2018fairness,biega2018equity,DBLP:conf/kdd/BeutelCDQWWHZHC19,Rastegarpanah:2019:FFF:3289600.3291002}. 

\paragraph{Popularity Bias in Recommenders}
Popularity bias is used as a running example in our experiments. It has been reported in many recommender systems \cite{Bellogin2017, DBLP:conf/flairs/AbdollahpouriBM19, channamsetty2017recommender}. However, the cause of popularity bias is not fully understood. One assumption is that popularity is an effective attribute to consider because users do prefer popular items \cite{steck2011item, cha2009analyzing, channamsetty2017recommender, jannach2013recommenders}. \citet{Canamares:2018:IFC:3209978.3210014} conducted a formal probabilistic analysis of the effectiveness of popularity-based recommendation approaches. \citet{steck2011item}
quantitatively studies the trade-off between item popularity and recommendation accuracy, and show with empirical evidence that even a small bias towards the long tail of the popularity distribution are considered to be valuable. 
On the opposite side, \citet{channamsetty2017recommender} instead found that user's preference towards popular items do not propagate to the recommendations they receive, so popularity bias in recommendations is not inherited from user preference. Moreover, \citet{cremonesi2010performance} discovered that popularity bias is dataset dependent. Empirically, on the MovieLens dataset, the effectiveness of recommending top rated or the most popular items changes drastically with the removal of a small number of top popular items.

\paragraph{Simulation studies}

Simulation has a long history in evaluating information retrieval and recommenders. \citet{cooper1973simulation} simulated documents and queries to understand how their characteristics influenced retrieval quantity. Metrics like expected search length~\citep{cooper1968expected}, discounted cumulative gain~\citep{jarvelin2002cumulated}, or rank-based precision~\citep{moffat2008rank}  use simple models of how users interact with ranked items to evaluate performance. Many works focus on making simulated user models more realistic~\citep[e.g.,][]{tague1980problems,mostafa2003simulation,diaz2009adaptation,carterette2011simulating}, though in this work we prefer to design simple users that will probe aspects of a recommender's behaviors (Section~\ref{sec:user_models}).

Simulations of multiple interactions with a recommender have been studied. \citet{szlavik2011diversity} studies six user choice models with different behavior in terms of tendency to accept recommendations and tendency to prefer highly rated trendy items through simulation on real and synthetic datasets. \citet{Jannach:2015:RRA:2852079.2852083} also run simulations of different recommender models to study their trajectories, but in that case studying the decreased diversity across user trajectories.
\citet{montaner2004evaluation} proposes a evaluation method for recommender systems through simulation, which progressively discover user profiles based on lists of evaluated items provided by simulated users. These studies all focus on evaluating recommender systems in terms of accuracy or diversity rather than analyzing the recommender's impact.

Simulation also has received increased attention as a means for testing reinforcement learning generally~\citep{brockman2016openai} and in recommender settings~\citep{ie2019recsim}. 

\citet{fairness_is_not_static} makes the case for analyzing the effects of decision making systems in simulation as an important part of responsible algorithm development.

\paragraph{Temporal analysis of recommender systems} Beyond the complexity in popularity bias shown in one-step recommendation, how popularity bias changes over time is even less predictable. For example, user preference may shift over time. \citet{jiang2019degenerate} studies the interaction between users and recommendation systems and proposes mitigation methods to slow down system degeneracy.
\citet{warlop2018fighting} discuss the effect of boredom and found that user satisfaction decreases as the recency and frequency of engagement increases. Also, as the number of observed items by each user increase during the process of simulation, the items available to be recommended at each step is different. This requires an understanding of the distribution of item popularity. \citet{cha2009analyzing} studies the popularity characteristics of user generated content systems and found that popularity can be modeled as a power-law distribution. \citet{fleder2009blockbuster} found that recommenders decrease diversity over time and creates a richer-get-richer effect for popular items; \citet{chaney2018algorithmic} shows similar pattern and points out that this homogeneity effect hinders user-item match and decreases recommendation utility.

\section{Simulation Framework} \label{sec:simulation_framework}
Simulation frameworks have been used for understanding RL \cite{broden2017bandit, zhao2017deep, brockman2016openai}, even for recommender systems \cite{ie2019recsim}, but less research has been done on how simulation can uncover behavioral patterns of recommender systems. As such, we offer a general framework for using simulations to evaluate recommender system performance.
We begin here by describing the simulation framework and the design decisions required to get meaningful results.

\subsection{Notation}
\label{sec:notation}
Consider a user that can interact with items $v \in \mathcal{V}$. The user's interaction at time $t$ has three aspects to it:

\begin{itemize}
    \item $s_t \subset \mathcal{V}$: The {\em slate}, i.e., items that were visible to the user when they made the choice.
    \item $c_t \in s_t$: The {\em choice}, i.e., which item the user chose to interact with.
    \item $r_t \in \mathbb{R}$: The {\em rating}, i.e., a measure of the user's satisfaction with the item.
\end{itemize}

A recommender system $\model_r$ is a function that maps a user history of interactions to a slate of $k$ item recommendations:
$$
\model_r: [(s_1, c_1, r_1),  (s_2, c_2, r_2), ...] \rightarrow \mathcal{V}^k.
$$

The repeated interactions of a user and the recommender give rise to a recursive generative model of user trajectories. Let  $\model_r$ be a recommender model, $\model_u$ be a model of user behavior, and $\tau_{:t}$ be a user's trajectory up to time $t$, i.e., 
$$\tau_{:t} = [(s_1, c_1, r_1),  ..., (s_{t-1}, c_{t-1}, r_{t-1})].$$

The next step is generated as:
\begin{align}
    s_t &= \model_r(\tau_{:t}) \\
    c_t, r_t &\sim \model_u(s_t) \label{eq:user_notation}
\end{align}

A user trajectory $\tau$ is a random variable drawn from a distribution parameterized by a length $T$, an initial history $\tau_0$, a user model $\model_u$, and a recommender model $\model_r$, i.e., $\tau \sim \mathcal{S}(\model_u, \model_r, \tau_0, T)$. 

An item $v$ may have other attributes associated with it, which we use $\rho(v)$ to denote generally. Examples of attributes that may be of interest could be $v$'s popularity or quality, e.g. $v$'s nutritional content in a food recommender. In experiments we will focus on popularity because it has been well-studied.

In attempting to use simulated samples $\tau$ to probe the impacts of the recommender $\model_r$, it is tempting to think that we need very realistic user models. Realistic modeling of users, however, is difficult (at least as hard as learning good recommenders), but we can use specifically designed user models to probe particular extreme cases that help us understand the range of behaviors a recommender can exhibit. We start by describing some standard recommender models in \S \ref{sec:api} and then the design of test users in more detail in \S \ref{sec:user_models}.

\subsection{Example Recommender Models}
\label{sec:api}

The recommender system $\model_r$ is initially {\em trained} on a data set $D$ that contains a set of users $\mathcal{U}$ and items $\mathcal{V}$ and their interactions.

\subsubsection{Matrix factorization}
\label{sec:mf_background}
For a matrix factorization model \citep{koren2009matrix}, the training involves learning a matrix of {\em item representations} $Q$ and {\em user representations} $P$, where each item and user is represented as a $d$-dimensional vector. Let $c_{i, j}$ be an indicator of {\em whether} user $i$ gave a rating to item $j$ in the training data, and $r_{i, j}$ be the rating that they gave. 
\begin{align}
\min_{P, Q} \sum_{i, j} c_{i, j}(p_i^T q_j - r_{i, j})^2 + \lambda \Omega(P, Q).
\label{eq:mf_train}
\end{align}
Here, $p_i$ is the $d$-dimensional representation for user $i$, $q_j$ is the $d$-dimensional representation for item $j$,
$\Omega$ is a regularizer of the parameters of the model and $\lambda$ is a cross-validated regularization strength.

At inference time, given a user's history, we first infer a personalized vector representation of the user and then provide a slate that maximizes their expected ratings from among items that they have not yet interacted with. The personalized vector $p$ and slate $s$ are computed as:

\begin{align}
\label{eq:mf_inference}
p &= \argmin_{p\prime} \sum_{v \in \mathcal{V}_{seen}} (p\prime^T q_v - r_{v})^2 \\
s &= \topk_{v \in \mathcal{V}_{unseen} } p^{T} q_v,
\end{align} where $\mathcal{V}_{seen}$ is the set of items that the user has interacted with,  $\mathcal{V}_{unseen}$ is the items that the user has never interacted with and $r_v$ is the rating that the user gave to $v \in \mathcal{V}_{seen}$.

\subsubsection{Recurrent neural networks}
\label{sec:rnn_background}
More recently, recurrent neural networks (RNNs) have also been used as recommender models \cite{hidasi2015session,rrn}. Here too, the ranking model infers a user's expected satisfaction with an item by the dot product between a user representation $p \in \mathbb{R}^d$ and an item representation $q \in \mathbb{R}^d$. However, here we consider the user representation to be the output of an RNN, $p = f_\theta(\tau_{:t})$ where $\theta$ are learned parameters of the RNN.

Training an RNN model consists of finding RNN parameters $\theta$ and $d$ dimensional item embeddings $Q$ that minimizes:
\begin{align}
\min_{\theta, Q} \sum_{u \in \mathcal{U}, t} (f_{\theta}(\tau_{u, :t}) \cdot q_{v_{u, t}} - r_{u, t})^2 + \lambda \Omega(\theta, Q).
\label{eq:rnn_train}
\end{align}
As in the matrix factorization equations, $\Omega$ is a regularizer of the parameters of the model and $\lambda$ is a regularization strength.

At inference time, given a user's history $\tau$, the recommended slate $s$ is then calculated as:
\begin{align}
s = \topk_{v \in \mathcal{V}_{unseen}} f_{\theta}(\tau) \cdot q_v.
\label{eq:rnn_inference}
\end{align}

\subsubsection{Design decisions}

\paragraph{Non-Repeating Recommendations}
Although the pure ML model may be inclined to recommend already observed items,
we only allow the recommender model to recommend from items that the user hasn't interacted with yet. We believe this is reasonable for two reasons. First, this aligns with the purpose of recommender systems, which is to explore unknown items that match user preference. Second, since previous interactions are used to train the recommenders, recommending from already observed items is equivalent to making inference on both training and testing sets, which biases towards training examples causing the recommender to be likely to more frequently recommend already observed items. With a small static dataset, we acknowledge the constraint in trajectory evaluation and comparison across users introduced by excluding observed items from recommendation, since the pool of unobserved items available to be recommended becomes dependent of user history.
However, this effect is negligible with a larger dataset and percentile definition of popularity.

\paragraph{No model retraining during a simulated trajectory but recommendations are still personalized} 
Both the matrix factorization and the RNN recommenders are continuously observing new data as a simulated user interacts sequentially with the recommender and their parameters could potentially be updated after each new interaction. However, fully online learning of a model's parameters (i.e., updating model parameters after \emph{every} new observation) is extremely rare in commercial recommender systems due to the difficulty of the infrastructure required. While model parameter updates (i.e., updating $Q$ in \eqref{eq:mf_train}, and $Q, \theta$ in \eqref{eq:rnn_train}) may take place relatively infrequently compared with the pace at which a user interact with the system, a user's trajectory $\tau$ {\em is} updated after every interaction. As a result, even though we hold the model parameters ($Q$ in \eqref{eq:mf_train}, and $Q, \theta$ in \eqref{eq:rnn_train}) constant, the user's representation (Equations \eqref{eq:mf_inference}, and $f_\theta(\tau)$ in \eqref{eq:rnn_inference}) is updated with every interaction -- even between model parameter retrainings.

\section{Designing test user models} 
\label{sec:user_models}

The user's behavior $\mathcal{M}_u$ naturally decomposes into two sub-models of user behaviors: the {\em selection} model $\mathcal{M}_{s}$ describes how a user chooses an item from a slate, and the {\em feedback} model $\mathcal{M}_{f}$ describes how a user rates an item after interacting with it. Thus, equation~\eqref{eq:user_notation} can be written in more detail as:
\begin{equation}
    c_t \sim \model_s(s_t), \hspace{3em} r_t  \sim \model_f(c_t)
\end{equation}

Both of these models can be considered methods of feedback of the user to the recommender. $\mathcal{M}_{s}$ affects for which items the recommender receives explicit feedback, whereas $\mathcal{M}_{f}$ controls the feedback itself. $\mathcal{M}_{s}$ is sometimes called the user's implicit feedback contrasted with $\mathcal{M}_f$ which is the user's explicit feedback.

As describe above, users are responsible for selecting an item from the set of recommendations provided during the simulation. Numerous models of user behavior in recommenders have been studied and used in designing recommenders  \cite{joachims2002optimizing,craswell2008experimental,kveton2015cascading,benson2016modeling}, but here we focus on designing test users summarized in Table~\ref{table:user_models} that will reveal different aspects of the recommender.

\subsection{Choice models}

One way of describing the effect of a system is to observe the trajectories of users who repeatedly follow its recommendations, and display no preference of their own. To test for this, we simulate different simple choice models related to the {\em rank} of an item in the slate provided by the recommender (i.e., the highest ranked item has a rank of 1, the second is 2, etc.). These selection models are: {\em lazy} (always accepts rank 1), {\em uniform} accepts an item uniformly at random, and {\em ranked} which accepts higher ranked items more often, with logarithmic decay. This model of user behavior is related to nDCG and follows~\citet{craswell2008experimental,joachims2002optimizing}.

To test the system's responsiveness to a user preference, we use {\em $\alpha$-preference}, which simulates users choosing items based on some attribute of an item $\rho(v)$ using a softmax distribution. The parameter $\alpha$ controls the strength of the user's preference for choosing items with that attribute, and negative values indicating an aversion.

\subsection{Feedback models}
While user selections (i.e. clicks) are important to many recommender systems, the user's subsequent interactions are also often important for the system to make good recommendations. For example, ads systems often consider conversions \cite{DBLP:conf/www/BarbieriSL16}, news recommenders may use dwell time \cite{lamba2019modeling} or reading time \cite{yi2014beyond}, and both product and movie recommendation systems can allow for explicit user ratings \cite{bennett2007netflix,linden2003amazon}.  This user feedback is important to the model's understanding of the user, and thus we also include various feedback models items to test the recommender system's responsiveness to this feedback. 

We consider a binary feedback mechanism (i.e., \{+1, -1\}) and two different types of feedback models: 1. {\em positive} feedback, which is always $+1$, and 2.  discretized attribute-based {\em $\beta$-preference} feedback, where the user's feedback depends on whether an item's attribute $\rho(v)$ exceeds a threshold $\rho_0$ and a sign $\beta \in \{-1, +1\}$ which indicates whether the user prefers content with high or low values of $\rho(v)$. 

\renewcommand{\arraystretch}{1.5}
\begin{table}
\begin{tabular}{|c|c|c|}
\hline
& {\bf Name} & {\bf Formula} \\
\hline
     \multirow{ 3}{*}{Choice} & Lazy &  $\model_s^{lazy}(v) =  \mathbb{1}[\text{Rank}(v) = 1]$ \\
     & Uniform&  $\model_s^{uniform}(v) =  \frac{1}{k}$ \\
     & Ranked &  $\model^{ranked}_s(v) \propto \frac{1}{\log(1 + \text{Rank}(v))}$ \\
     & $\alpha$-preference  & $\model_s^{\alpha-prefer}(v) \propto e^{\alpha\rho(v)}$ \\
     \hline
     \multirow{ 2}{*}{Feedback} & Positive & $\model_f^{positive}(v) =  +1 $ \\
     & $\beta$-preference & $\model_f^{\beta-prefer}(v) = \begin{cases} \beta & \rho(v) \geq \rho_0 \\
                 - \beta & \text{else}
\end{cases}$\\
\hline
\end{tabular}
     \caption{Summary of user selection and feedback models. \label{table:user_models}}
\end{table}
\renewcommand{\arraystretch}{1.0}

\subsection{Trajectory seeds}

The recommendations provided by the recommender also depend on the initial trajectory seed $\tau_0$ that the test user has at the start of their simulated interaction with the recommender. We consider both {\em random} seeds which consist of a single item uniformly chosen, as well as {\em real} seeds based on existing datasets. While the real seeds are more realistic, the random seeds allow for auditing of unlikely, but still possible, experiences that can arise for cold-start users.

\section{Empirical Case Studies} 
\label{sec:experiments}

We now explore what simulation-based audits can tell us in the context of three recommender systems.

As popularity bias has been studied extensively in MovieLens \cite{jannach2013recommenders, Canamares:2018:IFC:3209978.3210014}, we first use our simulation framework to explore trajectories created from matrix factorization over MovieLens \citep{harper2016movielens} in the context of popularity. Subsequently, we explore the trajectories framed from a large-scale, production 
RNN-based recommender \citep{hidasi2015session,rrn,DBLP:conf/wsdm/BeutelCJXLGC18,Chen:2019:TOC:3289600.3290999}. 

Throughout, we will consider {\em popularity} to be the item-attribute $\rho$ of interest and we use $\rho(\tau)$ to denote the value of that attribute for each item $v$ in the trajectory $\tau$.

$$
\rho(\tau) = [\rho(c) \text{ for } c \in \tau]
$$

When referring to the {\em slope} of a trajectory, we fit a linear regression to values of $\rho(\tau)$ with the $x$ axis being simulation steps and report the slope of that regression line.

\begin{figure}[t]
    \centering
    \begin{subfigure}[b]{0.23\textwidth}
        \includegraphics[width=\textwidth]{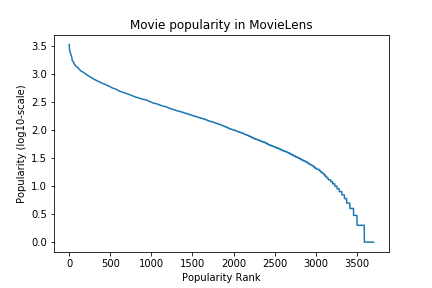}
        \caption{Item popularity}
        \label{fig:movie_pop}
    \end{subfigure}
    \begin{subfigure}[b]{0.23\textwidth}
        \includegraphics[width=\textwidth]{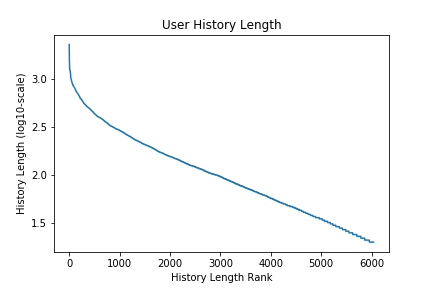}
        \caption{User history length}
        \label{fig:hist_len}
    \end{subfigure}
    \begin{subfigure}[b]{0.23\textwidth}
        \includegraphics[width=\textwidth]{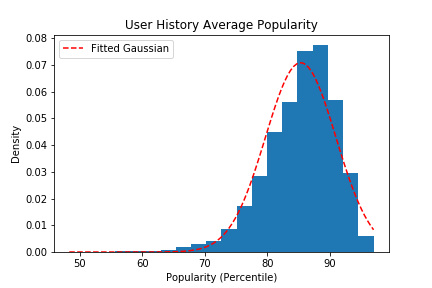}
        \caption{Popularity of rated items}
        \label{fig:hist_pop}
    \end{subfigure}
    \begin{subfigure}[b]{0.23\textwidth}
        \includegraphics[width=\textwidth]{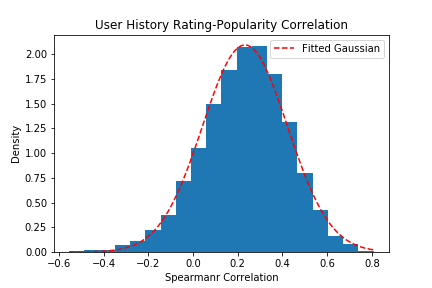}
        \caption{Corr(rating, popularity)}
        \label{fig:hist_corr}
    \end{subfigure}
    \caption{Distributional descriptions of the MovieLens 1M dataset.
    (a, b) Movie popularity and user history length follow power law distributions; (c) The average popularity of movies in users' watch histories is above the average item popularity; (d) Distribution of Spearman correlation between rating and popularity in user history, suggests that most users prefer popular items.}
    \label{fig:movielens_stats}
\end{figure}

\subsection{Matrix Factorization-based recommender}

We train a matrix factorization-based recommender (\S \ref{sec:mf_background}) on the MovieLens 1M dataset~\citep{harper2016movielens} which includes about 1 million ratings from 6040 users on 3952 movies. The distribution of item popularity is shown in Figure \ref{fig:movie_pop}. We seed each user with their initial trajectory $\tau_0$ from the dataset and simulate continuations of trajectories for another $T = 150$ steps.

\subsubsection{Trajectory shape without user preference}
\begin{figure*}[t]
    \centering
    \begin{subfigure}[b]{0.3\textwidth}
        \includegraphics[width=\textwidth]{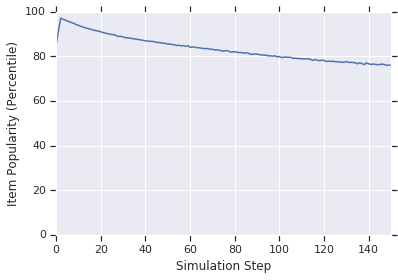}
        \caption{Average Trajectory}
        \label{fig:movielens_avg_traj}
    \end{subfigure}
    \begin{subfigure}[b]{0.3\textwidth}
        \includegraphics[width=\textwidth]{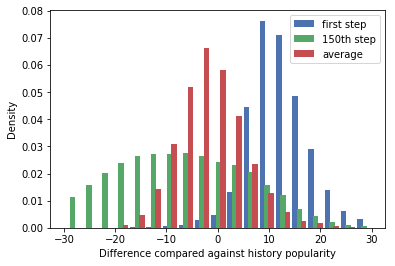}
        \caption{Diff in popularity}
        \label{fig:movielens_unbiased_traj_vs_history}
    \end{subfigure}
    \begin{subfigure}[b]{0.3\textwidth}
        \includegraphics[width=\textwidth]{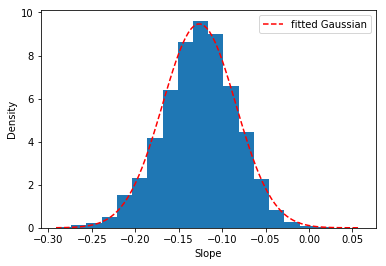}
        \caption{Slope}
        \label{fig:movielens_unbiased_slope}
    \end{subfigure}

    \caption{Trajectories for simulated test users no preferences interacting with a matrix factorization model. At first the recommended items have high popularity, but that decreases (and variance increases) over time.}
    \label{fig:unbiased}
\end{figure*}

We first evaluate the trajectories generated with uniform choice and  positive feedback models. As neither present any systematic user preference, any change in trajectory popularity cannot be attributed to the user. For each user, we measure the difference between the average popularity of items in the seed trajectory $\tau_0$, compared with the first, average, and last items in the simulated trajectory; the slope of $\rho(\tau)$, and the average value of $\rho(\tau)$ per step. The results are shown in Figure~\ref{fig:unbiased}.

As expected, the first step popularity is notably higher than the user history popularity (Figure~\ref{fig:movielens_unbiased_traj_vs_history}), echoing the popularity bias already reported in one-step recommendation research. However, in the subsequent steps, item popularity \emph{decreases} gradually since trajectory average popularity is lower than first step, and the 150th step popularity is lower than average. This is confirmed by the distribution of trajectory slope, as shown in Figure \ref{fig:movielens_unbiased_slope}, where we observe that almost all user trajectories have negative slope. We can see these statistics visually in Figure \ref{fig:movielens_avg_traj}, plotting the average trajectory under this model. Put together, this suggests a more nuanced story than much of the popularity bias literature: while we do observe a significant popularity bias in the recommender, over the course of multiple steps this decreases.

\subsubsection{Sensitivity to user preferences}
Next we would like to understand how user preferences, expressed through their choices and feedback, can affect trajectory shape.

\paragraph{Explicit user feedback}
We start with studying the effect of users' feedback. To isolate this effect, we use a uniform choice model so that there is no preference expressed by the choice model. We compare the always positive feedback model, with a preference based model using both positive preference (prefers popular content) and negative preference (prefers unpopular content). For the $\beta$-preference model, we binarize popularity using a threshold $\rho_0 = 1000$. Popularity values for each step of the trajectory are averaged over users.

\begin{figure}[t]
    \centering
    \includegraphics[width=0.7\columnwidth]{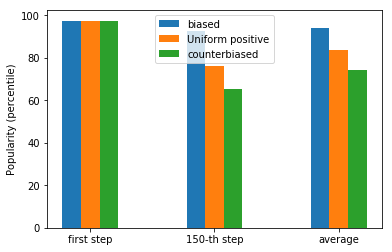}
    \caption{Explicit feedback: Popularity of items at the first step, last step, and averaged over the full simulation. Users give explicit feedback indicating either a preference for popular (biased), unpopular (counterbiased), or always positive. The recommender is responsive to the users' feedback.}
    \label{fig:compare_feedback_models}
\end{figure}

Figure~\ref{fig:compare_feedback_models} suggests that the matrix factorization model is responsive to explicit user feedback. On average, simulated users who express a preference for more popular content do receive it over time, and users express a preference for unpopular content receive it as well. Users who express no preference at all (always positive) are in between.

\paragraph{Selection preference}

\begin{figure}[t]
    \centering
    \begin{subfigure}[b]{0.22\textwidth}
        \includegraphics[width=\textwidth]{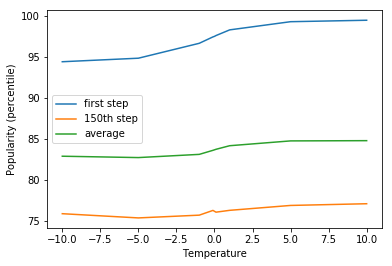}
        \caption{Popularity}
        \label{fig:traj_stats_vs_temp}
    \end{subfigure}
    \begin{subfigure}[b]{0.22\textwidth}
        \includegraphics[width=\textwidth]{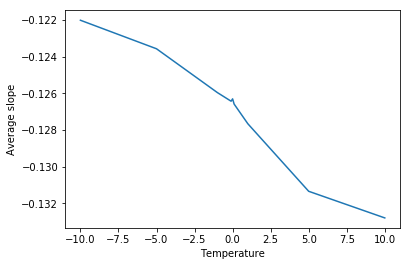}
        \caption{Slope}
        \label{fig:traj_slope_vs_temp}
    \end{subfigure}
    \caption{Selection preference: Popularity of items at the first step, last step, and averaged over the full simulation for simulated users with different degrees of selection preference for popular content. Users that prefer to select popular content consume more popular content overall, but have stronger downward slopes, suggesting that they are presented with less and less popular content as the simulation progresses.}
    \label{fig:compare_selection_ml}
\end{figure}

Next, we study the effect of user preference or bias in what they click on.
User choice introduces implicit bias in terms of whether popular items are more frequently observed and used to train the subsequent models. Here, we measure trajectory statistics with different values of $\alpha$ for $\alpha$-preference user choice models (from Table~\ref{table:user_models}). Average popularity values at different points in the trajectory are plotted in Figure \ref{fig:compare_selection_ml}. 

Having a stronger choice preference for popular items does increase the overall popularity of items that the a simulated user consumes over the course of their trajectory (Figure~\ref{fig:traj_stats_vs_temp}).  However, all of the trajectories slope {\em downwards}, suggesting that over time, the simulated users are being presented with less and less popular items. 

Notably, effect size of the selection bias is less than that of the feedback bias.

\subsection{Real-World Neural Recommenders}

\newcommand{\myparagraph}[1]{\textbf{#1:}\xspace}

We now turn toward understanding large-scale neural recommenders used in practice. This recommender system uses an RNN over the sequence of items consumed (\S~\ref{sec:rnn_background}), similar to \cite{hidasi2015session,rrn,DBLP:conf/wsdm/BeutelCJXLGC18,Chen:2019:TOC:3289600.3290999}. The model takes as input the sequence of items, timing, and other features, and predicts which item the user will consume next, with the loss weighted by measurements of user satisfaction. The model makes recommendations from a set of millions of items. Unless stated otherwise, the model produces a slate of size $k=10$. The popularity of an item is based on the number of times an item has been consumed historically by users of a real recommendation environment, but we report popularity based on the percentile of the item's popularity in the item vocabulary. This RNN model does not use explicit user feedback, so the user feedback model $\mathcal{M}_{f}$ is irrelevant (or alternatively, always positive). Features that the RNN expects other than the item chosen, such as timing, use standardized defaults.

\paragraph{Seeding trajectories}
For these experiments, we focus on the case of {\em fully synthetic} histories generated by simulated interactions with the recommender\footnote{The neural model can be interpreted as an auto-regressive model of the item the user will choose to interact with next, so the trajectories have realistic structure to them.}.  This gives us insight into the recommendation experience for new users and sheds light on the implications of the cold-start problem.  For each synthetic user, we seed the trace with a single random interaction from the item set which serves as the seed trajectory, $\tau_0$.

\begin{table*}[t]
    \centering
    \begin{tabular}{|l|l|r|r|r|}
    \hline
       & & Uniform & Ranked & Lazy \\ \hline\hline
        & 5\%ile & -15.9 & -14.4 & -12.0 \\\cline{2-5}
        First Step Increase & Mean & 33.5 & 33.8 & 37.5 \\\cline{2-5}
        & 95\%ile & 89.1 & 89.2 & 92.3 \\\hline\hline

        & 5\%ile & -0.425 & -0.4 & -0.513 \\\cline{2-5}
        Slope & Mean & -0.028 & -0.014 & -0.099 \\\cline{2-5}
        & 95\%ile & 0.443 & 0.478 & 0.394 \\\hline\hline
        \multicolumn{2}{|l|}{Percent with Positive Slope} & 41.1\% & 41.4\%  & 31.4\% \\\hline
        \multicolumn{2}{|l|}{Initial Step Popularity} & $49.6 \pm 29.2$ & $49.6 \pm 29.1$ & $49.6 \pm 29.1$ \\\hline
        \multicolumn{2}{|l|}{50th Step Popularity} & $82.8 \pm 20.0$ & $80.2 \pm 21.0$ & $80.2 \pm 21.0$ \\\hline
    \end{tabular}
    \caption{Trajectory statistics for simulated users with no preference. Item popularity values are presented as a percentiles.}
    \label{tab:traj_prod}
\end{table*}

\begin{table*}[t]
    \centering
    \begin{tabular}{|l|l|r|r|r|r|}
    \hline
    & & \multicolumn{4}{|c|}{Initial Item Popularity Percentile} \\
       & & 0-25th & 25th-50th & 50th-75th & 75th-100 \\ \hline\hline
        & 5\%ile          & 0.0 & 0 & -12.6 & -45.0\\\cline{2-6}
        First Step Increase & Mean & 66.1 & 45.9 & 20.1 & -1.53 \\\cline{2-6}
        & 95\%ile         & 95.6 & 70.9 & 44.6 & 19.8 \\\hline\hline

        & 5\%ile     & -0.419 & -0.388 & -0.414 & -0.471\\\cline{2-6}
        Slope & Mean & 0 & 0.008 & -0.007 & -0.067 \\\cline{2-6}
        & 95\%ile    & 0.536 & 0.481 & 0.394 & 0.332 \\\hline\hline
        \multicolumn{2}{|l|}{Percent with Positive Slope} & 44.6\% & 44.2\%  & 38.8\% & 36.3\%  \\\hline
        \multicolumn{2}{|l|}{Initial Step Popularity} & $13.8 \pm 7.3$  & $37.1 \pm 7.1$  & $62.5 \pm 7.0$ & $88.4 \pm  7.3$ \\\hline
        \multicolumn{2}{|l|}{50th Step Popularity} & $81.9 \pm 19.7$ & $82.4 \pm 20.2$ & $82.9 \pm 20.7$ & $84.0 \pm 19.6$ \\\hline
    \end{tabular}
    \caption{Trajectory statistics split by popularity of the user's seed item (for a Uniform user model).}
    \label{tab:traj_randomuser_popsplit}
\end{table*}

\begin{figure*}[t]
    \centering
    \begin{subfigure}[b]{0.3\textwidth}
        \includegraphics[width=\textwidth]{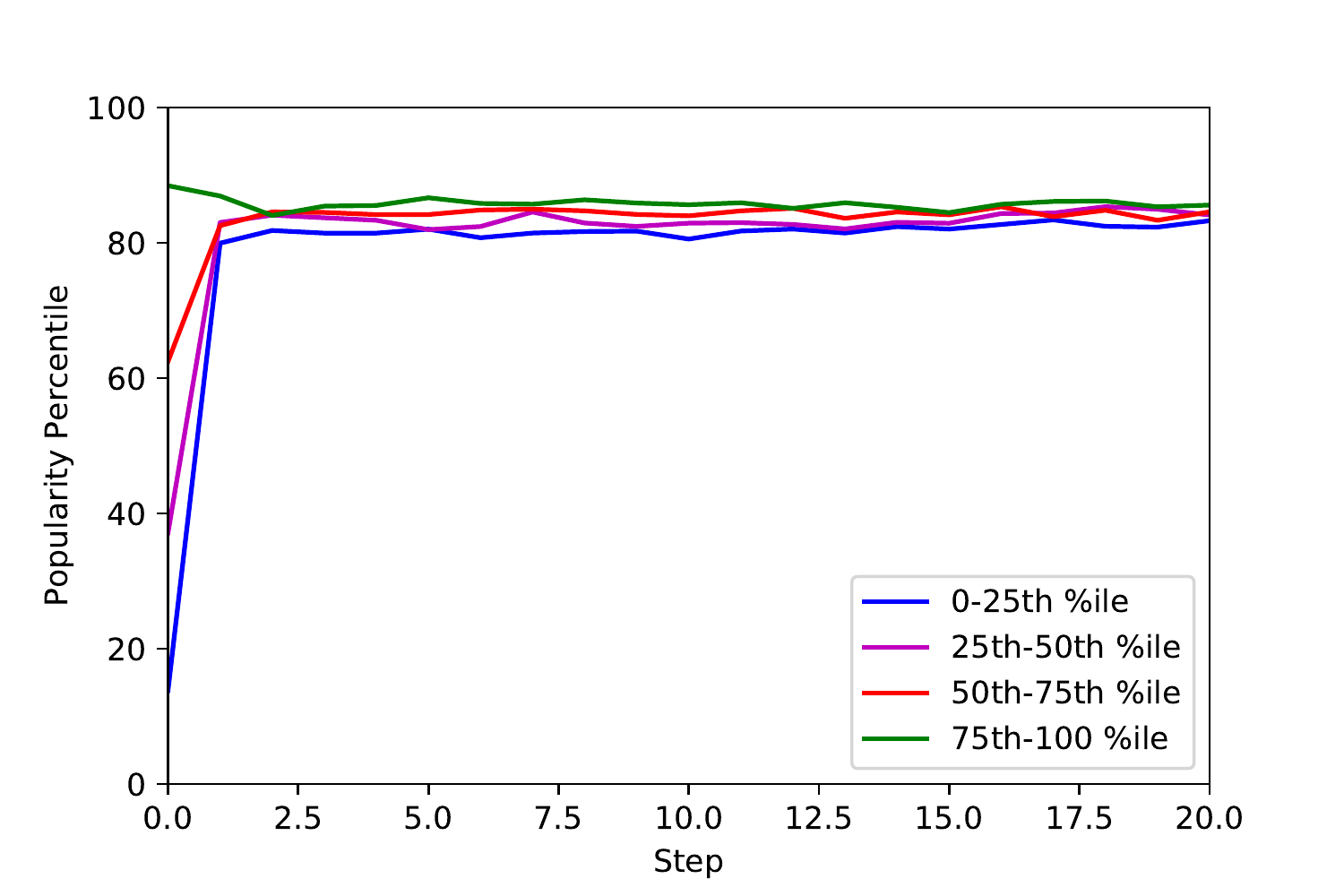}
        \caption{Uniform}
        \label{fig:rnn_random}
    \end{subfigure}
    \begin{subfigure}[b]{0.3\textwidth}
        \includegraphics[width=\textwidth]{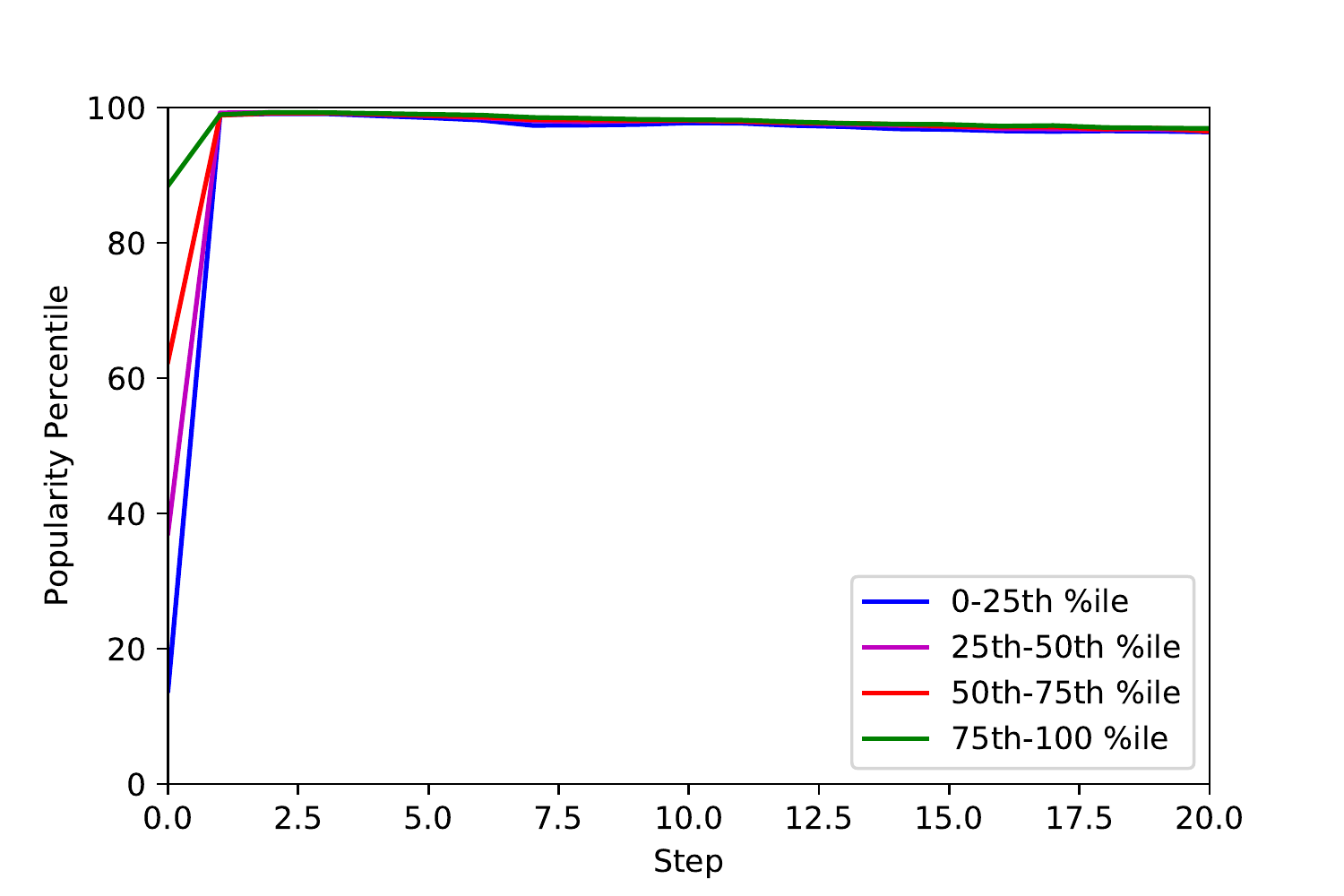}
        \caption{Popularity-Biased ($\alpha=50$)}
        \label{fig:rnn_pop}
    \end{subfigure}
    \begin{subfigure}[b]{0.3\textwidth}
        \includegraphics[width=\textwidth]{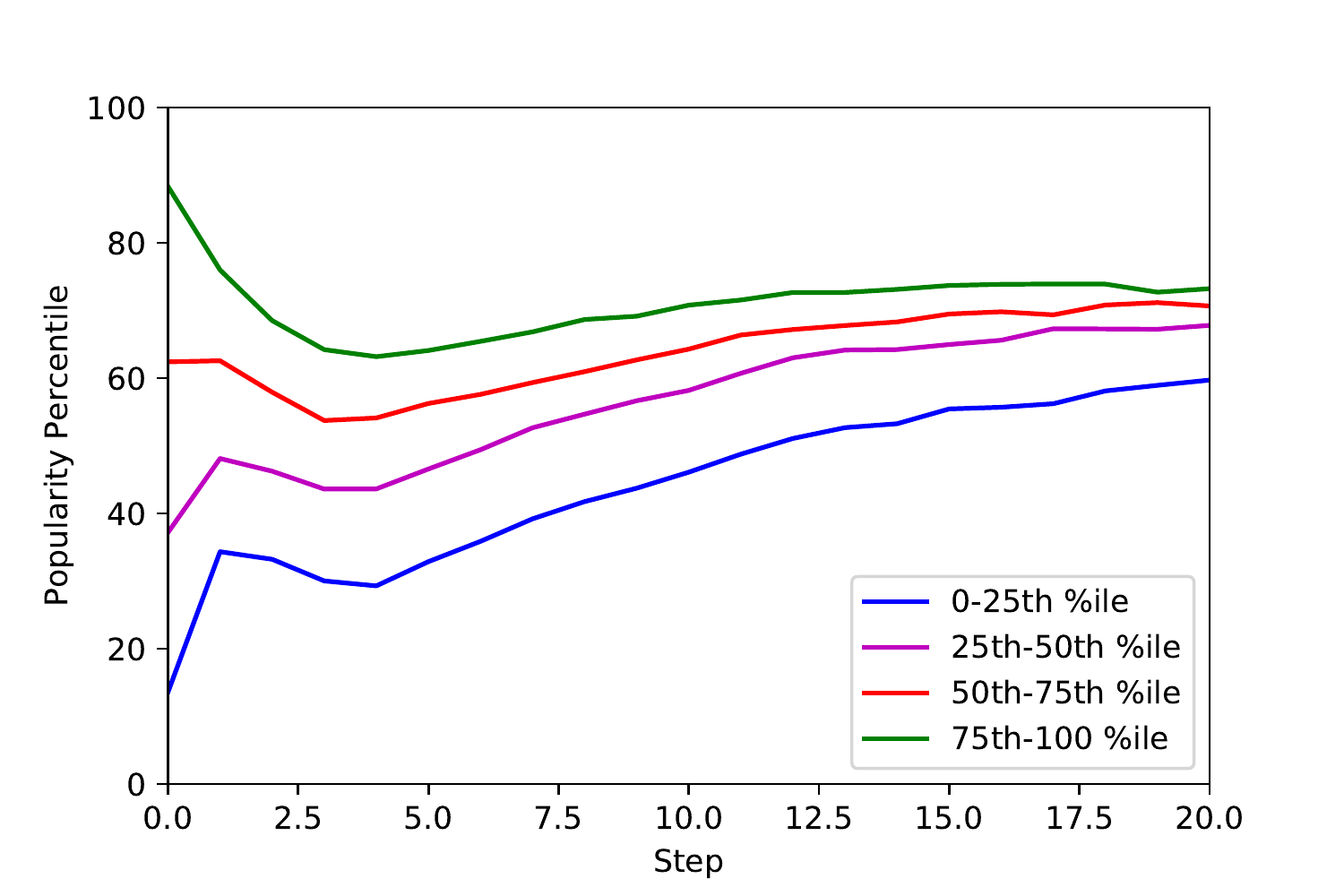}
        \caption{Anti-Popularity-Biased ($\alpha=-50$)}
        \label{fig:rnn_antipop}
    \end{subfigure}
    \caption{RNN simulated user trajectories on split by starting quartile over different types of user models. For popularity-averse simulated users, the resulting trajectories are sensitive to the popularity of the initial seed.}
    \label{fig:rnn_traj}
\end{figure*}

\subsubsection{Trajectory shape without user preference}
We begin with an analysis of trajectories for simulated users with no preferences. We simulate test users with choice models from \S \ref{sec:user_models}.

In Table \ref{tab:traj_prod} we present the RNN recommender interacting with three types of test users: uniform, ranked, and lazy (described previously in \S~\ref{sec:user_models}).  On average there is a significant increase in the popularity in even the first recommendation above the user's history (in this case, a random initial item). However, as in the matrix factorization case above, the long-term simulation yields a more nuanced picture of how popularity bias plays out.

First, we find that although the first step does show a significant increase in popularity, the average trend is a slight {\em decrease in popularity} over the course of the trajectory.  Second, there is high variability between users with the same (uniform) choice model. Despite the average slope being negative, over 40\% of the trajectories exhibit a positive slope. The highest 5\% of trajectories have slopes higher than 0.44.

\paragraph{Controlling for User History ($\tau_0$)}
Given the above results, we now consider how the initial item, which is not from the recommender system, effects the subsequent trajectories.  To do this, we split trajectories, based on the random user model, by the user's starting state, i.e., the first random item, into quartiles.  We plot the average trajectory for the four quartiles in Figure \ref{fig:rnn_random} and detail the statistics of the trajectories in Table \ref{tab:traj_randomuser_popsplit}.  Here we observe that nearly {\em independent of the starting item}, the first step moves all trajectories to significantly popular items (around the 80th percentile), but for users that start with a popular item, the recommender actually recommends a less popular item on average in the first step.  As we see in Table \ref{tab:traj_randomuser_popsplit}, for users that started with a less popular item, there is a slight positive slope on average and for users that started with a more popular item there is a slight negative slope, with almost 60\% of trajectories having a negative slope. Independent of the starting popularity, we find that the popularity after 50 steps is on average between the 80th percentile and the 85th percentile.
Considering the simulated user is simply choosing uniformly at random, these dynamics suggest that in some cases the model is either (a) reacting strongly to pure noise or (b) ignoring the information in the user's starting item and assuming a fairly popular item is a good prior assumption.

\begin{figure}[t]
    \centering
    \begin{subfigure}[b]{0.32\columnwidth}
        \includegraphics[width=\textwidth]{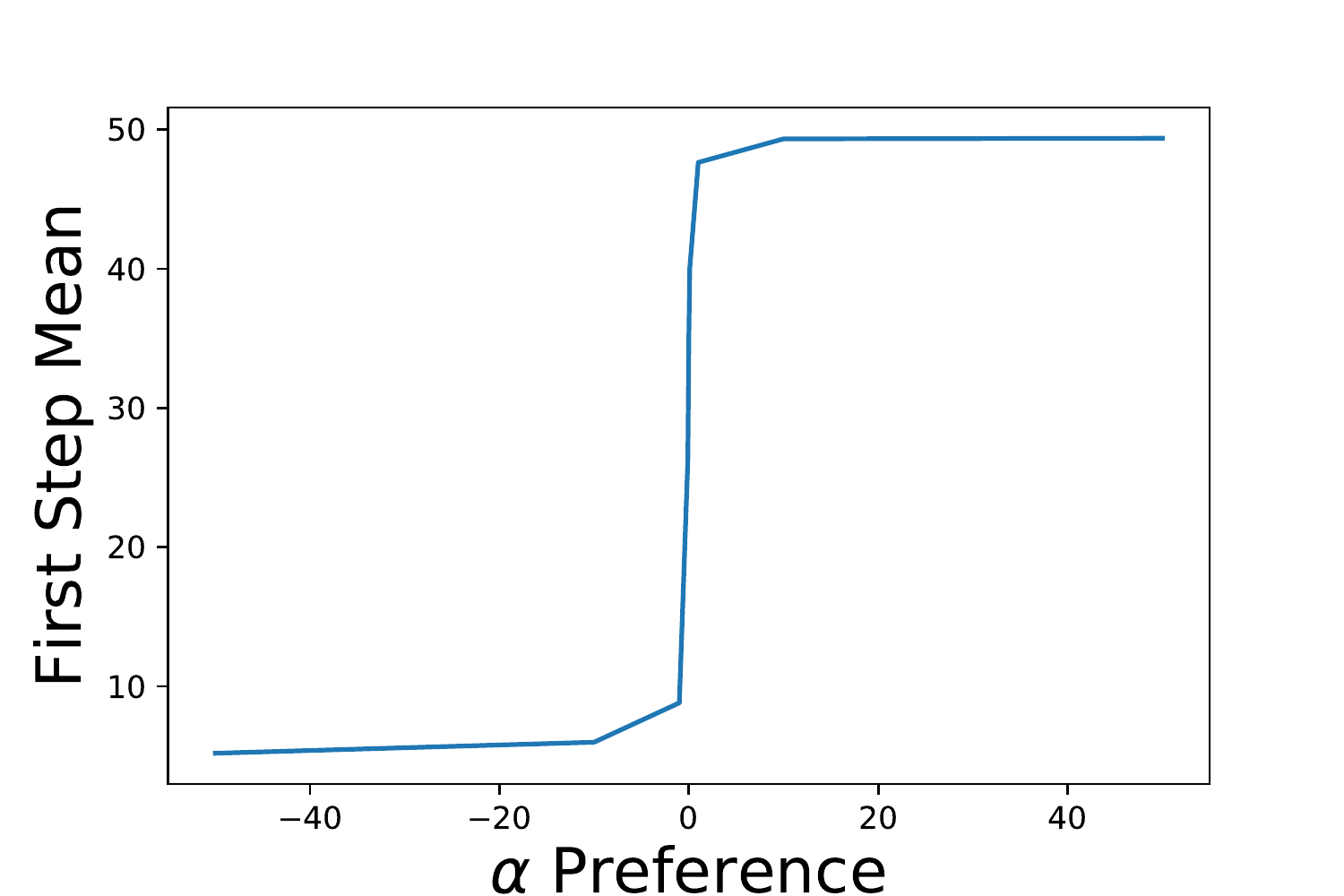}
        \caption{First Step}
    \end{subfigure}
    \begin{subfigure}[b]{0.32\columnwidth}
        \includegraphics[width=\textwidth]{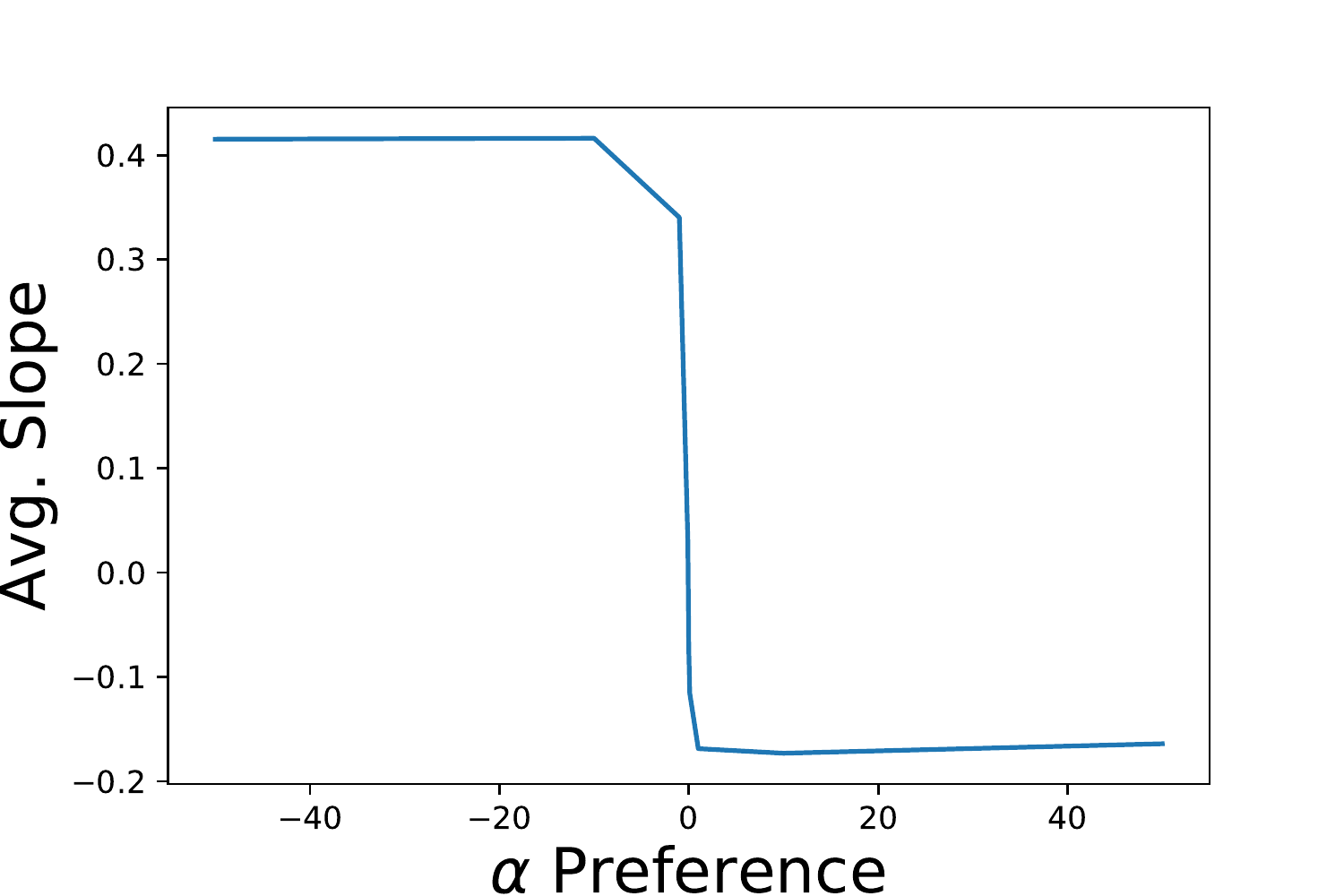}
        \caption{Slope}
    \end{subfigure}
    \begin{subfigure}[b]{0.32\columnwidth}
        \includegraphics[width=\textwidth]{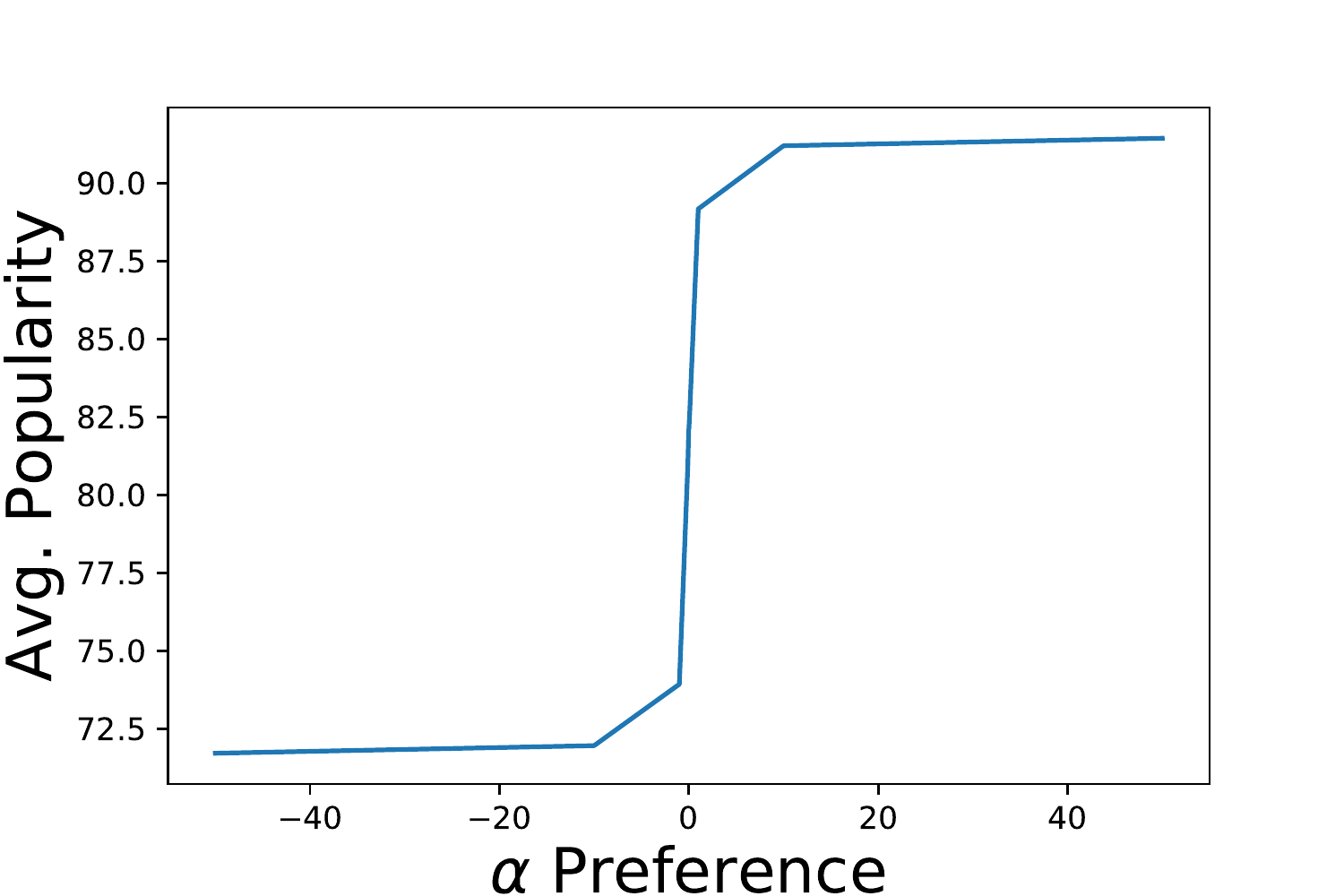}
        \caption{50th Step}
    \end{subfigure}
    \caption{Simulated users with varying degrees of preference for popular content ($\alpha$) interact with an RNN recommender. After the initial step, the trend is against the user's preference (b), moving users to a narrower range of item-popularity.}
    \label{fig:softmax_sensitivites}
\end{figure}

\paragraph{Responsiveness to user preference}
The above experiments use a purely random test user with no preferences. We now consider simulating test users with preferences using the $\alpha$-preference choice model in Table~\ref{table:user_models}.

In Figure \ref{fig:rnn_pop} and \ref{fig:rnn_antipop}, we plot the average trajectory for the cases of  $\alpha=\{50,-50\}$. We again split the trajectories by the starting item's popularity percentile. We see that for popularity-preferring users, the trajectories quickly converge to the most popular items, with little sensitivity to the starting item's popularity. For popularity-averse users, the popularity of videos in the trajectory does seem to be converging to a middling popularity, although after 20 steps there are significant differences depending on the popularity of the seed video. 

In Figure \ref{fig:softmax_sensitivites}, we plot the first-step popularity bias, the slope of the trajectory, and the average end item popularity for user models with $\alpha$ preference values of $\{-50, -10, -1, -0.1, -0.01, 0.001, 0.1, 1, 10, 50 \}$.  There we observe clear patterns: the user model has a significant effect on the first step, with user-bias for popular items resulting in a very large increase in popularity and user-bias against popular items resulting in a slight increase in popularity.  However, we find that following this initial step, the average slope matches those in Figure \ref{fig:rnn_traj}, with popularity-preferring users showing a negative slope and popularity-averse users showing a positive slope. This results in the trajectories, after 50 steps, ending in a narrower popularity range than at the initial step. However, despite this, we do see evidence that the RNN system is somewhat responsive to user preference over time.

\section{Discussion}
As we see in all of the experiments above, recommender systems show non-trivial temporal dynamics even with random user behavior, and the combined effects of system biases and user behaviors are more complicated than a single-step analysis can show. We believe that this simulation provides an opportunity to uncover and test these temporal dynamics.

These tools and our empirical studies inspire numerous follow-up questions.  The analysis in this paper provides a more complete picture of \emph{what} is happening in recommender systems.  However, further research is needed to understand \emph{why} these patterns emerge from recommender systems.  For example, to what degree are these patterns the result of the training data, the model architecture, or training procedure (as in policy learning)?  Further, once we move past the single-step notion of bias, it becomes even less clear what types of trajectories a recommender \emph{should} create.  For example, is popular content a reasonable assumption for users with otherwise minimal and noisy preferences?  And if recommenders don't create the trajectories that we believe are best, \emph{how} can we train models to achieve a different user experience?  Research is needed to develop training methods that provide the flexibility to make these decisions.  Finally, while we focus our empirical analysis on popularity biases, the framework is general to other item attributes, and as such further research is needed to uncover other biases and dynamics at play in recommenders.

Building a more nuanced understanding of recommender systems is a necessary first step in pursuing these questions, and we believe these simulation-based evaluations provide a tractable approach to understanding the temporal dynamics in these complex systems.

\section{Conclusion}
In this paper, we propose a simulation framework that goes beyond one-step recommendation and incorporates the interaction between user preferences and system effects, to better understand recommender system biases over time. We design test users to isolate the recommender's default long-term behaviors as well as its responsiveness to explicit and implicit user preferences. Finally, we run thorough experiments on both a matrix-factorization-based recommender on MovieLens and an RNN-based recommender in  a large-scale production recommender system to understand their behaviour and how popularity bias manifests in repeated interactions between the user and the recommender.

\begin{acks}
Thanks to Konstantina Christakopoulou, Fernando Diaz, and anonymous reviewers for feedback on the publication.
\end{acks}

\bibliographystyle{ACM-Reference-Format}
\bibliography{references}

\end{document}